%% file: main.tex
\documentclass[letterpaper]{article}
\usepackage{ijcai16}
\usepackage{times}
\usepackage{helvet}
\usepackage{courier}
\usepackage{color}
\usepackage{graphicx}
\usepackage{float}
\usepackage{subfig}
\usepackage{array}
\usepackage{epsfig}
\usepackage{amsmath}
\usepackage{bm}
\usepackage{caption}
\title{Entity Embedding-based Anomaly Detection for Heterogeneous Categorical Events}
\author{Ting Chen$^1$\thanks{Part of the work is done during first author's internship at NEC Labs America.}, Lu-An Tang$^2$, Yizhou Sun$^1$, Zhengzhang Chen$^2$, Kai Zhang$^2$ \\
$^1$Northeastern University, $^2$NEC Labs America  \\ \{tingchen, yzsun\}@ccs.neu.edu, \{ltang, zchen, kzhang\}@nec-labs.com}

\DeclareMathOperator*{\argmax}{arg\,max}

\begin{document}
	
\newcommand{\comment}[1]{\textcolor{red}{[\textit{#1}]}} 
\newcommand{\nop}[1]{}
\newcommand{\modelname}[0]{APE}

\maketitle

\input{abstract}
\input{introduction}
\input{problem}
\input{model}

\input{exp}

\input{relatedwork}

\input{conclusion}

\section*{Acknowledgement}

We would like to thank Zhichun Li, Haifeng Chen, Guofei Jiang, and Jiawei Han for some helpful discussions. This work is partially supported by NSF CAREER \#1453800, Northeastern TIER 1, and Yahoo! ACE Award.

\bibliographystyle{named}
\bibliography{main}

\end{document}

%% file: abstract.tex
\begin{abstract}

Anomaly detection plays an important role in modern data-driven security applications, such as detecting suspicious access to a socket from a process. In many cases, such events can be described as a collection of categorical values that are considered as entities of different types, which we call heterogeneous categorical events. Due to the lack of intrinsic distance measures among entities, and the exponentially large event space, most existing work relies heavily on heuristics to calculate abnormal scores for events. Different from previous work, we propose a principled and unified probabilistic model $\modelname{}$ (\underline{A}nomaly detection via \underline{P}robabilistic pairwise interaction and \underline{E}ntity embedding) that directly models the likelihood of events. 
In this model, we embed entities into a common latent space using their observed co-occurrence in different events. More specifically, we first model the compatibility of each pair of entities according to their embeddings. Then we utilize the weighted pairwise interactions of different entity types to define the event probability.
Using Noise-Contrastive Estimation with ``context-dependent" noise distribution, our model can be learned efficiently regardless of the large event space. Experimental results on real enterprise surveillance data show that our methods can accurately detect abnormal events compared to other state-of-the-art abnormal detection techniques.

\end{abstract}

%% file: introduction.tex
\section{Introduction}


With increasing amount of data collected from everywhere, such as computer systems, transaction activities, social networks, it becomes more and more important for people to understand the underlying regularity of the data, and to spot the unexpected or abnormal instances \cite{chandola2009anomaly}. Centered around this goal, anomaly detection plays a very important role in many security related applications, such as securing enterprise network by detecting abnormal connectivities, and so on.


However, the problem has not been satisfyingly addressed yet. Many traditional anomaly detection methods focus on either numerical data or supervised settings \cite{chandola2009anomaly}. When it comes to unsupervised anomaly detection in heterogeneous categorical events data, i.e., events containing a collection of categorical values that are considered as entities of different types, there is less existing work \cite{das2007detecting,das2008anomaly,tong2008fast,akoglu2012fast}.

The heterogeneous categorical event data are ubiquitous, such as events of process interactions in computer systems, where each data point is an event that involves heterogeneous types of attributes/entities: time, user, source process, destination process, and so on. In order to detect abnormal events that deviate from the regular patterns, a common approach is to build a model that can capture the underlying factors/regularities of data. However, events with multiple heterogeneous entities are difficult to model in a systematic and unified framework due to two major challenges: (1) the lack of intrinsic distance measures among entities and events, and (2) the exponentially large event space.

Consider that in real computer systems, given two users with ids of 1 and 10, we almost know nothing about their distance/similarity without other information. In addition to the lack of intrinsic distance measure, the exponentially large event space is also an issue. For example, a heterogeneous categorical event, in real systems, can involve more than ten types of entities. If each entity type has more than one hundred possible choices of entities the overall event space will be as large as $100^{10}$, which is prohibitively large and makes it challenging to model regularities.

Due to these two difficulties, most existing work relies heavily on heuristics to quantify the normal/abnormal scores for events \cite{das2007detecting,das2008anomaly,tong2008fast,akoglu2012fast}. However, a more systematic and accurate model is in demand as the vastly emerging of big complicated data in important applications.


To tackle the aforementioned challenges, we propose a probabilistic model that directly models the event likelihood. We first embed entities into a common latent space where distance among entities can be naturally defined. Then to access the compatibility of entities in the event, we quantify their pairwise interactions by the dot product of the embedding vectors. Finally the weighted sum of interactions is used to define the probability of the event.


Compared to traditional methods, the proposed method has several advantages: (1) by modeling the likelihood of event based on entity embeddings, the proposed model can produce normal/abnormal score in a principled and unified framework; (2) by modeling weighted pairwise interaction instead of all possible interactions, the model is less susceptible to over-fitting, and can provide better interpretability; and (3) the proposed model can be learned efficiently by Noise-Contrastive Estimation with ``context-dependent" noise distribution regardless of large event space. Empirical studies on real-world enterprise surveillance data show that by applying our method we can detect unknown abnormal events accurately.

%% file: problem.tex
\section{Problem Statement}
\label{sec:problem}

Here we introduce some notations and define the problem.

\textbf{Heterogeneous Categorical Event.} A heterogeneous categorical event ${e}=(a_{1}, \cdots, a_{m})$ is a record contains $m$ different categorical attributes, and the $i$-th attribute value $a_i$ denotes an entity from the type $A_i$. In the computer process interaction network, an event is a record involving entities of types such as the user, time\footnote{Although time is continuous value, it can be chunked into segments of different granularities, such as day and hour, which then can be viewed as entities.}, source/destination process and folder. In the following, we will call it event for short.

\begin{figure}[th]
\begin{center}
\epsfig{file=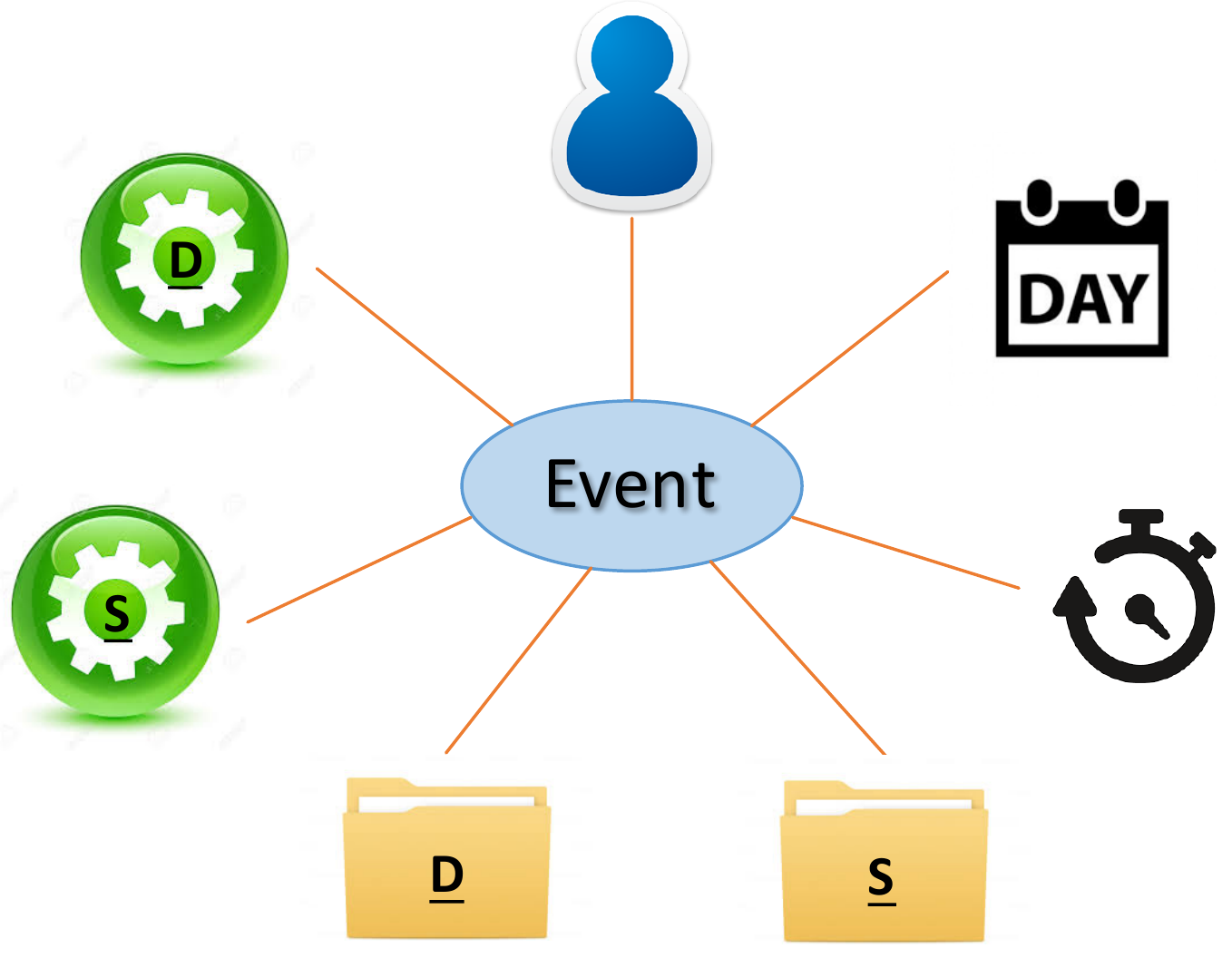, height=3.8cm}
\end{center}
\caption{\label{fig:schema} The heterogeneous network view of categorical events. Node types include event, user, day, hour, source/destination process and folder.}
\end{figure}

By treating the categorical attributes of an event as entities/nodes, we can also view categorical events as a heterogeneous network of multiple node types \cite{sun2012mining}. In the computer process interaction example, the network schema is shown in Figure \ref{fig:schema}, where event acts as a super node connecting other nodes of different types.

\textbf{Problem: abnormal event detection.} Given a set of training events $D =\{e_1, \cdots, e_n\}$, by assuming that most events in $D$ are normal, the problem is to learn a model $M$, so that when a new event $e_{n+1}$ comes, the model $M$ can accurately predict whether the event is abnormal or not.

%% file: model.tex
\section{The Proposed Model}

In this section, we introduce the motivation and technical details about the proposed model.

\subsection{Motivations}

We directly model the event likelihood as it indicates how likely an event should occur according to the data. An event with unusual low likelihood is naturally abnormal. To achieve this, we need to deal with the two main challenges as mentioned before: (1) the lack of intrinsic distance measures among entities and events, and (2) the exponentially large event space.

To overcome the lack of intrinsic distance measures among entities, we embed entities into a common latent space where their semantic can be preserved. More specifically, each entity, such as a user, or a process in computer systems, is represented as a $d$-dimensional vector and will be automatically learned from the data. In the embedding space, the distance of entities can be naturally computed by distance/similarity measures in the space, such as Euclidean distances, vector dot product, and so on. Compared with other distance/similarity metrics defined on sets, such as Jaccard similarity, the embedding method is more flexible and it has nice property such as transitivity \cite{kai2015sdm}.

To alleviate the large event space issue and enable efficient model learning, we come up with two strategies: (1) at the model level, instead of modeling all possible interactions among entities, we only consider pairwise interaction that reflects the strength of co-occurrences of entities \cite{rendle2010factorization}; and (2) at the learning level, we propose using noise-contrastive estimation \cite{gutmann2010noise} with ``context-dependent" noise distribution.

The pairwise interaction is intuitive/interpretable, efficient to compute, and less susceptible to over-fitting. Consider the following anomaly example we may encounter in real scenarios:

\begin{itemize}
\item A maintenance program is usually triggered at midnight, but suddenly it is trigged during the day.
\item A user usually connect to servers with low privilege, but suddenly it tries to access some server with high privilege.
\end{itemize}

In these examples, abnormal behaviors occur as a result of the unusual pairwise interaction among entities (process and time in the first example, and user and machine in the second example). 

\subsection{The probabilistic model for event}

We model the probability of a single event ${e} = \{a_1, \cdots, a_m\}$ in event space $\Omega$ using the following parametric form:

\begin{equation}
\label{eq:p}
P_\theta({e}) = \frac{\exp\bigg(S_\theta({e})\bigg)}{\sum_{{e}' \in \Omega}\exp\bigg(S_\theta({e}')\bigg)}
\end{equation}
Where $\theta$ is the set of parameters, $S_\theta({e})$ is the scoring function for a given event ${e}$ that quantifies its normality. We instantiate the scoring function by pairwise interactions among embedded entities:

\begin{equation}
\label{eq:score}
S_\theta({e}) = \sum_{i,j: 1\le i < j\le m}{w}_{ij} ( {v}_{a_i} \cdot {v}_{a_j})
\end{equation}
Where ${v}_{a_i}$ is the embedding vector for entity $a_i$, and the dot product between a pair of entity embeddings $v_{a_i}$ and $v_{a_j}$ encodes the compatibility of two entities co-occur in a single event. $w_{ij}$ is the weight for pairwise interaction between entity types $A_i$ and $A_j$, and it is non-negative constrained, i.e. $\forall i,j,  {w}_{ij}\ge 0$. Different pairs of entity types can have different importances, interaction among some pairs of entity types are very regular and important, e.g. user and machine, while others may be less regular and important, e.g. day and hour. Using $w_{ij}$, the model can automatically learn the importances of different pairwise interactions. Finally $\theta =\{w, v\}$ denotes all parameters used in the model.

\begin{figure}[!t]
\begin{center}
\epsfig{file=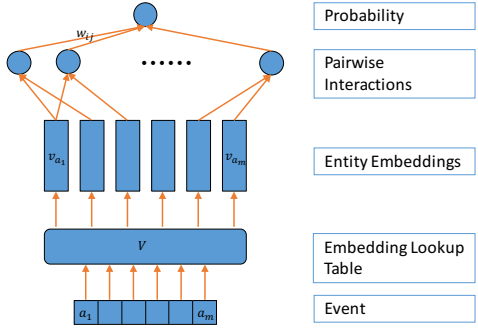,height=5cm}
\end{center}

\caption{\label{fig:architecture} The framework of proposed method.}

\end{figure}

Our model APE, which is abbreviated for \underline{A}nomaly detection via \underline{P}robabilistic pairwise interaction and \underline{E}ntity embedding, is summarized in Figure \ref{fig:architecture}.

The learning problem is to optimize the following maximum likelihood objective over events in the training data $D$:

\begin{equation}
\label{eq:obj}
\argmax_{\theta} \sum_{{e} \in D} \log P_\theta({e})
\end{equation}

To solve the optimization problem, the major challenge is that the denominator in Eq. \ref{eq:p} sums over all possible event configurations, which is prohibitively large ($O(\exp m)$). To address this challenging issue, we propose using Noise Contrastive Estimation.

\subsection{Learning via noise-contrastive estimation}

Noise-Contrastive Estimation (NCE) has been introduced in \cite{gutmann2010noise} for density estimation, and applied to estimate language model \cite{mnih2012fast}, and word embedding \cite{mnih2013learning,mikolov2013efficient,mikolov2013distributed}. The basic idea of NCE is to reduce the problem of density estimation to binary classification, which is to discriminate samples from data distribution $P_d(e)$ and some artificial known noise distribution $P_n(e)$ (the selection of $P_n$ will be explained later). In another word, the samples fed to the \modelname{} model can come from real training data set or being generated artificially, and the model is trained to classify them a posteriori.

Assuming generated noise/negative samples are $k$ times more frequent than observed data samples, the posterior probability of an event $e$ came from data distribution is $P(D = 1|e, \theta)\footnote{Since we want to fit the model distribution to data distribution, we use $P_\theta$ in place of $P_d$.} = {P_\theta(e)}/{(P_\theta(e) + k P_n(e))}$. To fit the objective in Eq. \ref{eq:obj}, we maximize the expectation of $\log P(D|e, \theta)$ under the mixture of data and noise/negative samples \cite{gutmann2010noise,mnih2012fast}. This leads to the following new objective function:

\begin{equation}
\label{eq:obj_nce}
\begin{split}
J(\theta) =& E_{e \sim {P}_d} \bigg[\log \frac{P_\theta(e)}{P_\theta(e) + kP_n(e)}\bigg] + \\
k& E_{e \sim P_n} \bigg[\log \frac{kP_n(e)}{P_\theta(e) + kP_n(e)}\bigg] 
\end{split}
\end{equation}

However, in this new objective function, the model distribution $P_\theta(e)$ is still too expensive to evaluate. NCE sidesteps this difficulty by avoiding explicit normalization and treating normalization constant as a parameter. This leads to $P_{\theta}(e) = P_{\theta^0} (e) \exp(c)$, where $\theta=\{\theta^0, c\}$, and $c$ is the original log-partition function as a single parameter, and is learned to normalize the whole distribution. Now we can re-write the event probability function in Eq. \ref{eq:p} as follows:

\begin{equation}
\label{eq:pnew2}
\begin{split}
P_\theta(e) = \exp \bigg(\sum_{i,j: 1\le i < j\le m} w_{ij} (v_{a_i} \cdot v_{a_j}) + c\bigg)
\end{split}
\end{equation}

To optimize the objective E.q. \ref{eq:obj_nce} given the training data $D$, we replace $P_d$ with $\tilde{P}_d$ (the empirical data distribution), and since the \modelname{} model is differentiable, stochastic gradient descent is used: for each observed training event $e$, first sample $k$ noise/negative samples $\{e'\}$ according to the known noise distribution $P_n$, and then update parameters according to the gradients of the following objective function (which is derived from Eq. \ref{eq:obj_nce} on given $e, \{e'\}$ samples):

\begin{equation}
\label{eq:obj_sgd}
\begin{split}
 & \log\sigma\bigg(\log P_\theta(e) - \log k P_n(e)\bigg) + \\
 & \sum_{e'} \log\sigma\bigg(-\log P_\theta(e') + \log k P_n(e') \bigg) 
\end{split}
\end{equation}
Here $\sigma(x) = 1/(1+\exp(-x))$ is the sigmoid function.

The complexity of our algorithm is $O(Nkm^2d)$, where $N$ is the number of total observed events it is trained on, $k$ is number of negative examples drawn for each observed event, $m$ is the number of entity type, and $d$ is the embedding dimension. The complexity indicates that the \modelname{} model can be learned efficiently regardless of the $O(\exp{m})$ large event space.

\subsection{``Context-dependent" noise distribution}

To apply NCE, as shown in Eq. \ref{eq:obj_sgd}, we need to draw negative samples from some known noise distribution $P_n$. Intuitively, the noise distribution should be close to the data distribution, otherwise the discriminating task would be too easy and the model cannot learn much structure from the data \cite{gutmann2010noise}. Note that, different from previous work (such as language modeling or word embedding \cite{mnih2012fast,mikolov2013efficient}) that utilizes NCE, where each negative sample only involves one word/entity. Each event, in our case, involves multiple entities of different types.

One straight-forward choice of noise distribution is \textit{``context-independent"} noise distribution, where a negative event is drawn independently and does not depend on the observed event. One can sample a negative event according to some factorized distribution on event space, i.e. $P_n^{factorized}(e) = p_{A_1}(a_1) \cdots p_{A_i}(a_i)$. Here $p_{A_i}(a_i)$ is the probability of choosing entity $a_i$ of the type $A_i$, which can be specified uniformly or computed by counting unigram in data. In this work we stick to unigram as it is reported better \cite{mnih2012fast,mikolov2013efficient}.

Although the ``context-independent" noise distribution is easy to evaluate. Due to the large event space, this noise distribution would be very different from data distribution, which will lead to poor model learning.

Here we propose a new \textit{``context-dependent"} noise distribution where negative sampling is dependent on its context (i.e. the observed event). The procedure is, for each observed event $e$, we first uniformly sample an entity type $A_i$, and then sample a new entity $a'_i \sim p_{A_i}(a'_i)$ to replace $a_i$ and form a new negative sample $e'$. As we only modify one entity in the observed event, the noise distribution will be close to data distribution, thus can lead to better model learning. However, by utilizing the new ``context-dependent" noise generation, it becomes very hard to compute the exact noise probability $P_n(e)$. Therefore, we use an approximation instead as follows.

For a given observed ``context" event $e$, we define the ``context-dependent" noise distribution for sampled event $e'$ as $P^c_n(e'|e)$. Since $e'$ is sampled by randomly replacing one of the entity $a_i$ with $a'_i$ of the same $A_i$ type, the conditional probability $P^c_n(e'|e) = P_{A_i}(a'_i)/m$ (here we assume $A_i$ is chosen uniformly). Considering the large event space, it is unlikely that event $e'$ is generated from observed events other than $e$, so we can approximate the noise distribution with $P_n(e') \approx P^c_n(e'|e) P_d(e)$. Furthermore, as $P_d(e)$ is usually small for most events, we simply set it to some constant $l$, which leads to the final noise distribution term (which is used in E.q. \ref{eq:obj_sgd}):
$$\log k P_n(e') \approx \log P_{A_i}(a'_i) + z,$$ where $z=\log kl/m$ is a constant term. Although we do not know the exact value of $z$, we let $z=0$ when plugging the approximated $\log kP_n(e')$ into Eq. \ref{eq:obj_sgd}. We find that ignoring $z$ will only lead to a constant shift of learned parameter $c$. Since $c$ is just the global normalization term, it will not affect the relative normal/abnormal scores of different events.

To compute $P_n(e)$ for an observed event $e$, since we do not know which entity is replaced as in the negative event case, we will use the expectation as follows: $$\log k P_n(e) \approx \sum_i \frac{1}{m} \log P_{A_i}(a_i) + z.$$ And again the $z$ will be ignored when plugging into Eq. \ref{eq:obj_sgd}.

%% file: exp.tex
\section{Experiments}

In this section, we evaluate the proposed method using real surveillance data collected in an enterprise system during a two-week period.

\subsection{Data Sets}

One of the main application scenarios of anomaly detection is to detect abnormal activity in surveillance data collected from computer systems. Hence, in our experiments, a two-week period of activity data of an enterprise computer system is used. The collected surveillance data include two types of events, which are viewed as two separate data sets.

\textbf{P2P.} Process to process event data set. Each event of this type contains the system activity of a process interacting with another process, the time and user id of the event are also recorded. P2P events are among the most important system activities since modern operating systems are based on processes. 

\textbf{P2I.} Process to Internet Socket event data set. Each event of this type contains the system activity of a process sending or receiving Internet connections to/from other machine at destination ports, the time and user id of the event are recorded as well. We only consider the P2I events among the enterprise system since we focus on inside enterprise activities.

The entity types and their number of entities for both data sets are summarized in Table \ref{tab:entity_types}.

\begin{table}[]
\centering
\small
\caption{Entity types in data sets.}

\label{tab:entity_types}
\begin{tabular}{|c|p{7.1cm}|}
\hline
Data & Types of entity and their arities                                                             \\ \hline \hline
P2P       & day (7), hour (24), uid (361), src proc (778), dst proc (1752), src folder (255), dst folder (415)                  \\ \hline
P2I       & day (7), hour (24), src ip (59), dst ip (184), dst port (283), proc (91), proc folder (70), uid (162), connect type (3) \\ \hline
\end{tabular}

\end{table}

We do not have the ground-truth labels for collected events, however, it is assumed that majority of events are normal. In order to evaluate anomaly detection task, similar to  \cite{das2007detecting,das2008anomaly,akoglu2012fast}, we create some artificial anomalies, and ask the algorithms to detect them. The artificial anomaly events are generated as follows: for each event in the test data, we select $c$ of its entities (we consider $c=\{1,2,3\}$ in following experiments), randomly replace them with other entities of the same type, and make sure the new generated events do not occur in both training and test data sets, so that they can be considered more abnormal than observed events. 

We split the two-week data into two of one-weeks. The events in the first week are used as training set\footnote{With randomly selected portion as validation set for selection of hyper-parameters.}, and \textit{new} events that only appeared in the second week are used as test sets. The statistics of observed events are summarized in Table \ref{tab:data_stat}.

\begin{table}[]
\centering
\small
\caption{Statistics of the collected two-week events.}

\label{tab:data_stat}
\begin{tabular}{|l|l|l|l|}
\hline
Data  & \# week 1 & \# week 2 & \# week 2 new \\ \hline\hline
P2P       & 95,434             & 107,619           & 53,478 (49.69\%)       \\ \hline
P2I       & 1,316,357          & 1,330,376         & 498,029 (37.44\%)       \\ \hline
\end{tabular}

\end{table}

\subsection{Comparing methods and settings}

We compare the following state-of-the-art methods for abnormal event detection.

\textbf{Condition.} This method is proposed in \cite{das2007detecting}. For each test event, it computes the conditional scores for all pairs of dependent and mutually exclusive subsets having up to $k$ attributes, and combine the scores with a heuristic algorithm. The conditional score is calculated based on statistics of events in the training set, and reflect dependencies between two given attribute sets of an event. 

\textbf{CompreX.} This method is proposed in \cite{akoglu2012fast}. It utilizes a compression technique to encode training data and learns a set of code tables that summarize patterns. When a new event comes, it first encodes it using existing code tables, and then the number of bits used in encoding is treated as abnormal score for the event.

\textbf{\modelname{}.} This is the proposed method. Noted that we use the negative of its likelihood output as the abnormal score.

\textbf{\modelname{} (no weight).} This method is the same as \modelname{}, except that instead of learning $w_{ij}$, we simply set $\forall i,j, w_{ij}=1$, i.e. it is APE without automatic weights learning on pairwise interactions. All types of interactions are weighted equally. 

For the (hyper-)parameter settings, we use part of the training data as validation set to tune (hyper-)parameters. For Condition, we set $k=1, \alpha=1, \beta=0.5$. For CompreX, we adopt their implementation, and since it is parameter free, we do not need to tune any parameters. For both \modelname{} and \modelname{} (no weight), the following setting is used: the embedding is randomly initialized, and dimension is set to 10; for each observed training event, we draw 3 negative samples for each of the entity type, which accounts for a total of $3m$ negative samples per training instance; we also use a mini-batch of size 128 for speed up stochastic gradient descent, and 5-10 epochs are general enough for convergence.

\subsection{Evaluation Metrics}

Since all methods listed above produce abnormal scores instead of binary labels, and there is no fixed threshold, thus metrics for binary labels such as accuracy are not suitable for measuring the performance. Similar to \cite{das2007detecting,akoglu2012fast}, we adopt ROC curves (Receiver Operating Characteristic curves) and PRC (Precision Recall curves) for evaluation. Both of these two curves reflect the quality of predicted scores according to their true labels at different threshold levels. A detailed discussion about the two metrics can be found in \cite{davis2006relationship}. To get a quantitative measurements, the AUC (area under curve) of both ROC and PRC are utilized.

\subsection{Results for abnormal event detection}

\begin{table*}[!tp]
\centering
\small
\caption{Performance of abnormal event detection. Values left to slash are AUC of ROC, and ones on the right are average precision. The last two rows ($^*$ marked) are averaged over 3 smaller ($1\%$) test samples due to long runtime of CompreX.}
\label{tab:p2p_p2i_abnormal_event}
\begin{tabular}{|c|l|l|l|l|l|l|}
\hline
\multicolumn{1}{|l|}{} & \multicolumn{3}{c|}{P2P}                                                       & \multicolumn{3}{c|}{P2I}                                                       \\ \hline
Models                 & \multicolumn{1}{c|}{c=1} & \multicolumn{1}{c|}{c=2} & \multicolumn{1}{c|}{c=3} & \multicolumn{1}{c|}{c=1} & \multicolumn{1}{c|}{c=2} & \multicolumn{1}{c|}{c=3} \\ \hline
Condition              & 0.6296 / 0.6777          & 0.6795 / 0.7321          & 0.7137 / 0.7672          & 0.7733 / 0.7127          & 0.8300 / 0.7688          & 0.8699 / 0.8165          \\ \hline
APE (no weight)        & 0.8797 / 0.8404          & 0.9377 / 0.9072          & 0.9688 / 0.9449          & 0.8912 / 0.8784          & 0.9412 / 0.9398          & 0.9665 / 0.9671          \\ \hline
APE                    & \textbf{0.8995 / 0.8845} & \textbf{0.9540 / 0.9378} & \textbf{0.9779 / 0.9639} & \textbf{0.9267 / 0.9383} & \textbf{0.9669 / 0.9717} & \textbf{0.9838 / 0.9861} \\ \hline \hline
CompreX$^*$               & 0.8230 / 0.7683          & 0.8208 / 0.7566          & 0.8390 / 0.7978          & 0.7749 / 0.8391          & 0.7834 / 0.8525          & 0.7832 / 0.8497          \\ \hline
APE$^*$                  & \textbf{0.9003 / 0.8892} & \textbf{0.9589 / 0.9394} & \textbf{0.9732 / 0.9616} & \textbf{0.9291 / 0.9411} & \textbf{0.9656 / 0.9729} & \textbf{0.9829 / 0.9854} \\ \hline
\end{tabular}
\end{table*}

\begin{figure}[!tp]
\centering
  \begin{tabular}{@{}c@{}}
    \includegraphics[width=0.95\linewidth,height=110pt]{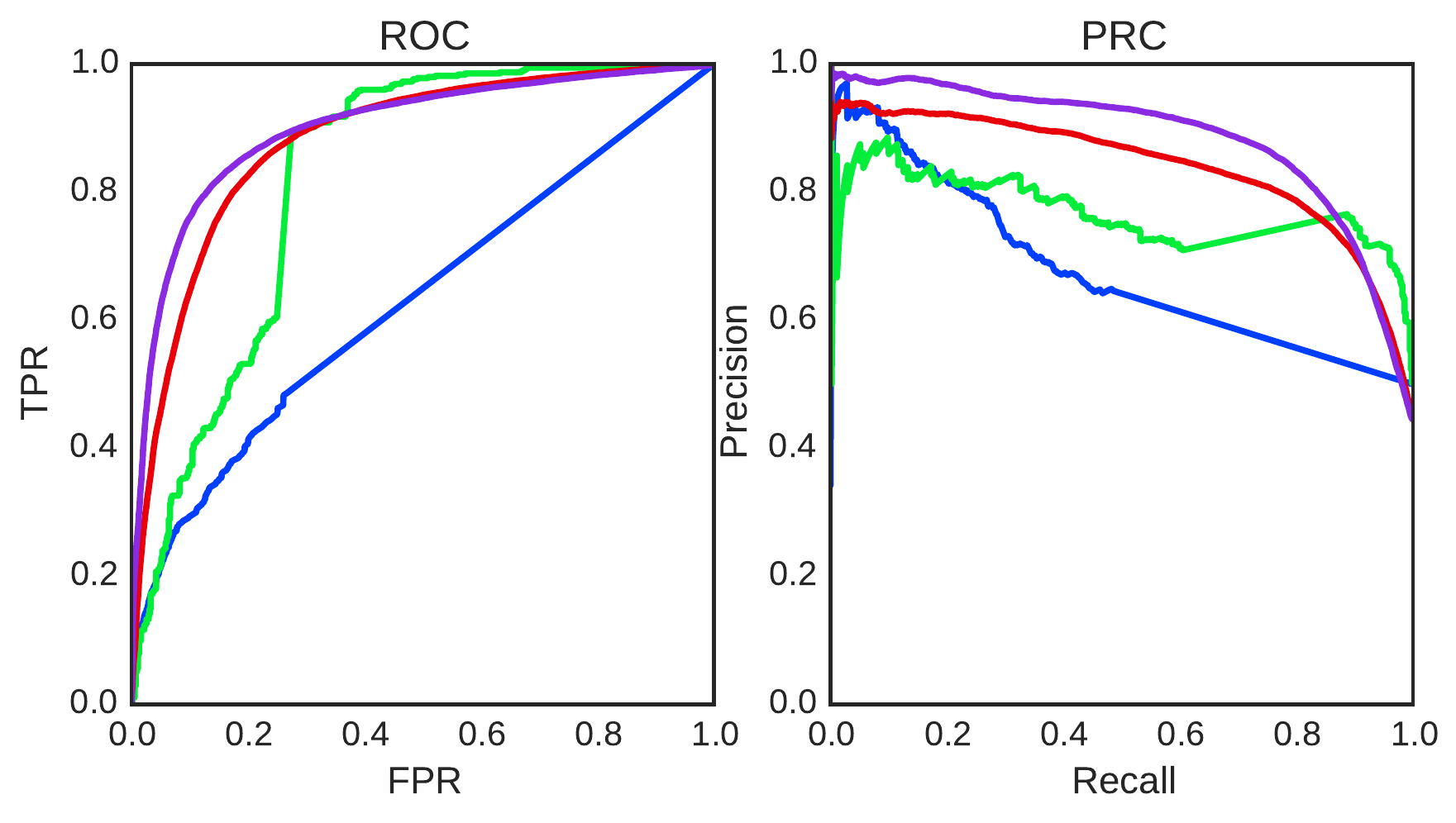} \\[\abovecaptionskip]
    \small (a) P2P abnormal event detection.
  \end{tabular}

    \begin{tabular}{@{}c@{}}
      \includegraphics[width=0.95\linewidth,height=110pt]{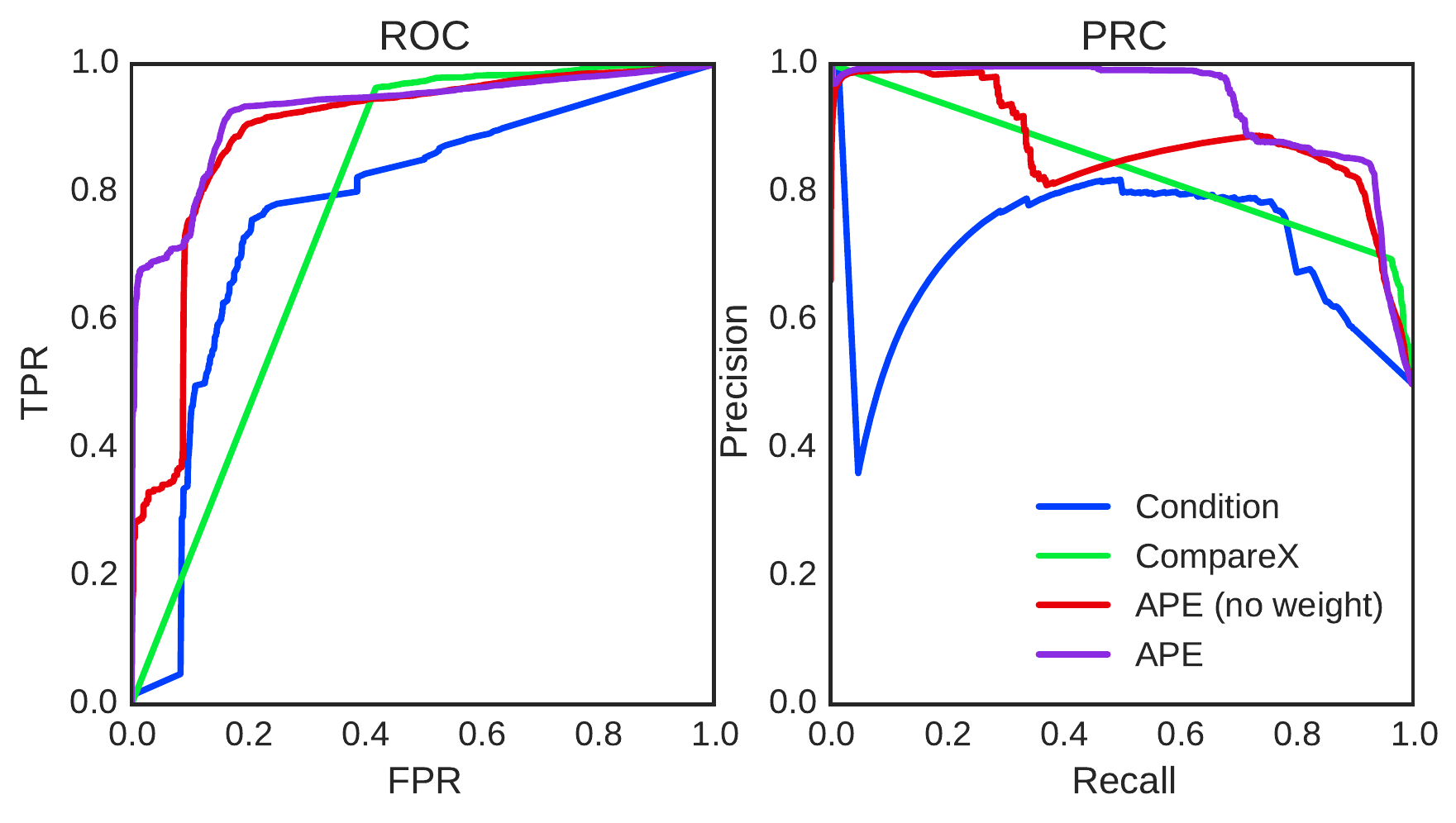} \\[\abovecaptionskip]
      \small (b) P2I abnormal event detection.
    \end{tabular}
    \caption{\label{fig:p2p_p2i_event_curves} Receiver operating characteristic curves and precision recall curves for abnormal event detections.}

\end{figure}

Table \ref{tab:p2p_p2i_abnormal_event} shows the AUC of ROC and PRC of different methods on P2P and P2I data sets. Note the last two rows in Table \ref{tab:p2p_p2i_abnormal_event} are mean scores averaged over three sampled smaller test sets, due to the slowness of CompreX at test time (which can takes hundreds of hours to finish on the half million sized P2I events). Figure \ref{fig:p2p_p2i_event_curves} shows both ROC curves and PR curves for all methods using test set with entity replacement $c=1$ (for $c=2, 3$, results are similar thus not shown).

From the results we can see, on different $c$ number of entity replacement, our method consistently outperforms both Condition and CompreX significantly. When comparing \modelname{} with \modelname{} (no weight), we see that by considering weights and learning them automatically, the detection results can be further improved.

\begin{figure}[h]
\begin{center}
\epsfig{file=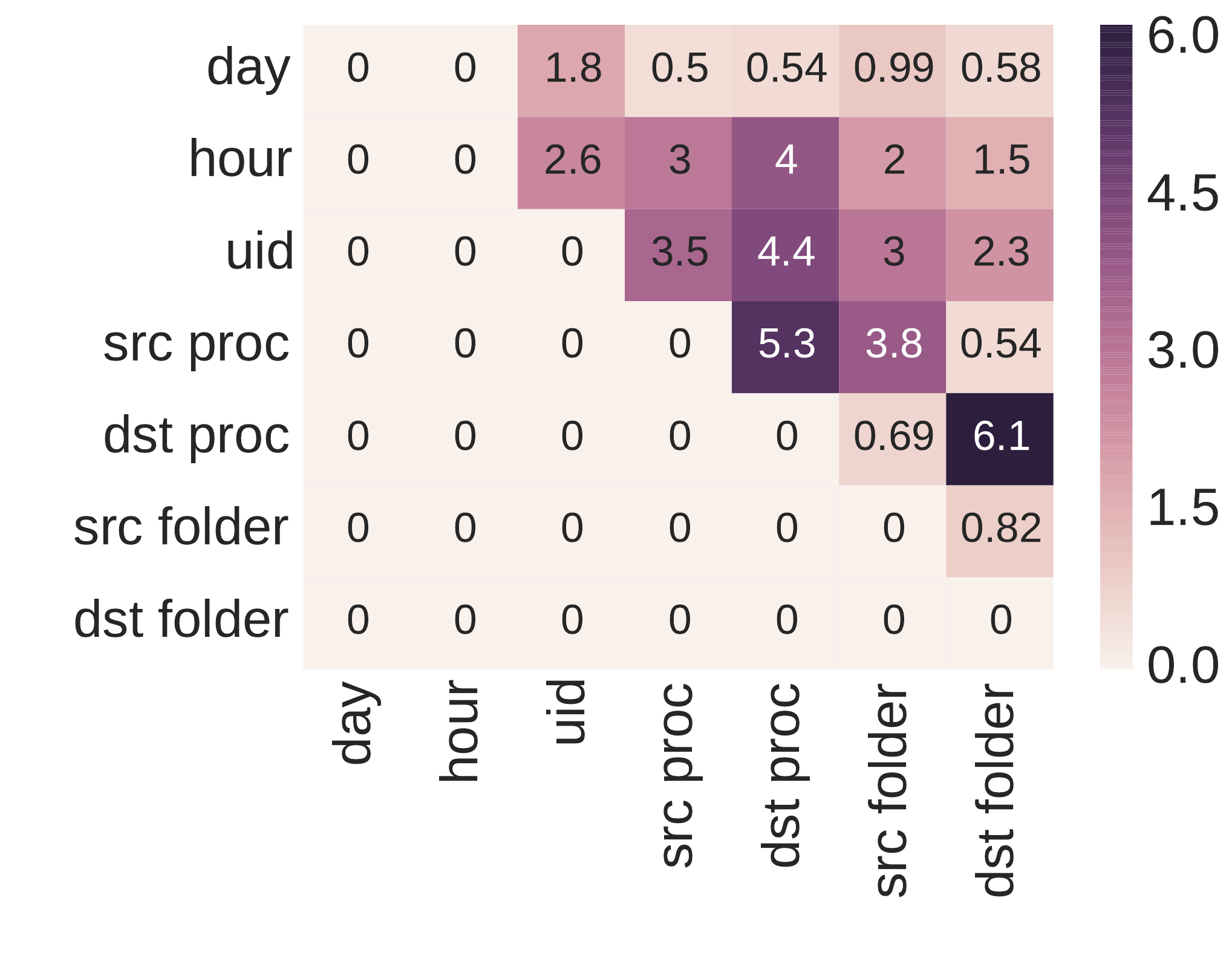,height=4.3cm}
\end{center}
\caption{\label{fig:weight_p2p} Pairwise weights learned for P2P events.}
\end{figure}

\begin{figure}[h]
\begin{center}
\epsfig{file=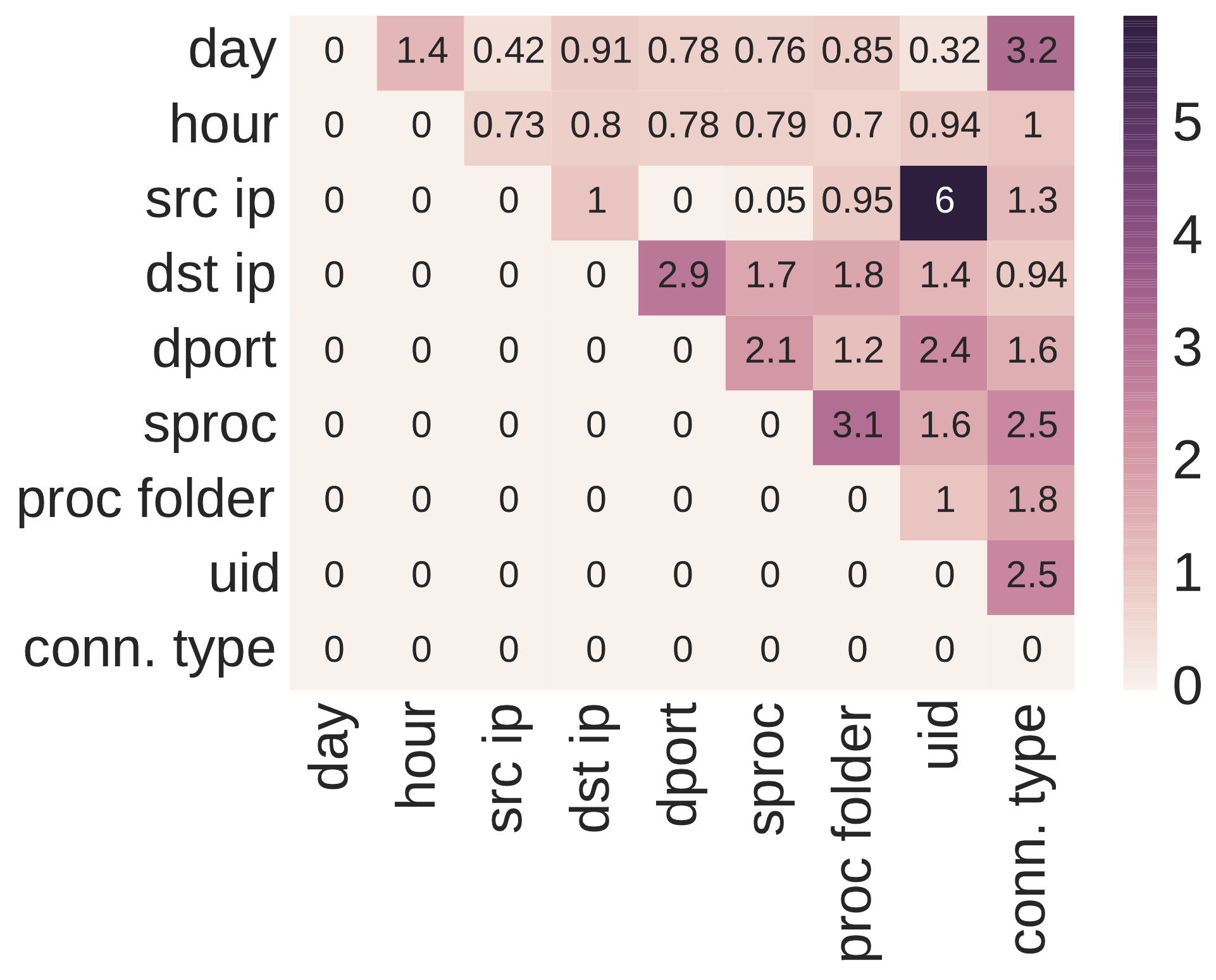,height=4.3cm}
\end{center}
\caption{\label{fig:weight_p2i} Pairwise weights learned for P2I events.}
\end{figure}

The learned weight matrix $W$ for P2P and P2I events can be found in Figure \ref{fig:weight_p2p} and \ref{fig:weight_p2i}, respectively. The matrix is upper-triangulated since the pairwise interaction is symmetric and model only among different type of entities. From the weights, we can see the importance of different types of interactions in the data sets. For example, in P2P events, the weight for interaction between day and hour is insignificant; while the weight for interaction between source process and destination process is large, indicating they are highly dependent and capture the regularity of P2P events.

Table \ref{tab:abevent_case} shows some detected abnormal events (we only highlight the pairs of entities that have the particular low comparability score). In the first event, the interaction between process bash and its folder is irregular and results in small likelihood; in the second event, the abnormality is caused by a main user (who usually active during the work hour) involved in the event on 1 a.m.; in the third example, the process ssh connects to an unexpected port 80 and thus raising the alarm.

\begin{table}[!tp]
\centering
\small
\caption{Detected abnormal events examples.}
\label{tab:abevent_case}
\begin{tabular}{|c|c|}
\hline
{ Data } & { Abnormal event}                         \\ \hline\hline
P2P            & ..., src proc: bash, src folder: /home/, ...              \\ \hline
P2P            & ..., uid: 9 (some main user), hour: 1, ...                       \\ \hline
P2I            & ..., proc: ssh, dst port: 80, ...                   \\ \hline
\end{tabular}
\end{table}

\subsection{Results for different noise distributions}

Table \ref{tab:noise_dist} shows performances under different choices of noise distribution. Results shown are collected from test set with $c=1$ (for $c=2,3$, the results are similar thus not shown), and using the same number of  training events. 

First we compare the ``context-independent" noise distribution (first row) and the proposed ``context-dependent" noise distribution (third row), clearly the ``context-dependent" one performs significantly better. This confirms that by using the proposed ``context-dependent" noise distribution, the \modelname{} model can learn much more effectively given the same amount of resources.

We also compare the importance of the approximated noise probability term $\log kP_n(e)$ in Eq. \ref{eq:obj_sgd}. Simply ignore the term by setting it to zero (second row) (as similarity used in \cite{mikolov2013efficient,mikolov2013distributed}) results in much worse performances compared to our proposed approximated one.

\begin{table}[h]
\small
\centering
\caption{Average precision under different choice of noise distributions.}
\label{tab:noise_dist}
\begin{tabular}{|c|c|c|}
\hline
     Noise distribution       & P2P             & P2I             \\ \hline\hline
Context-independent  & 0.8463 & 0.7534 \\ \hline
Context-dependent, $\log kP_n(e)=0$   & 0.8176 & 0.7868 \\ \hline
Context-dependent, $\log kP_n(e)=\text{appx}$ & 0.8845 & 0.9383 \\ \hline
\end{tabular}
\end{table}

\begin{figure}[]
\begin{center}
\epsfig{file=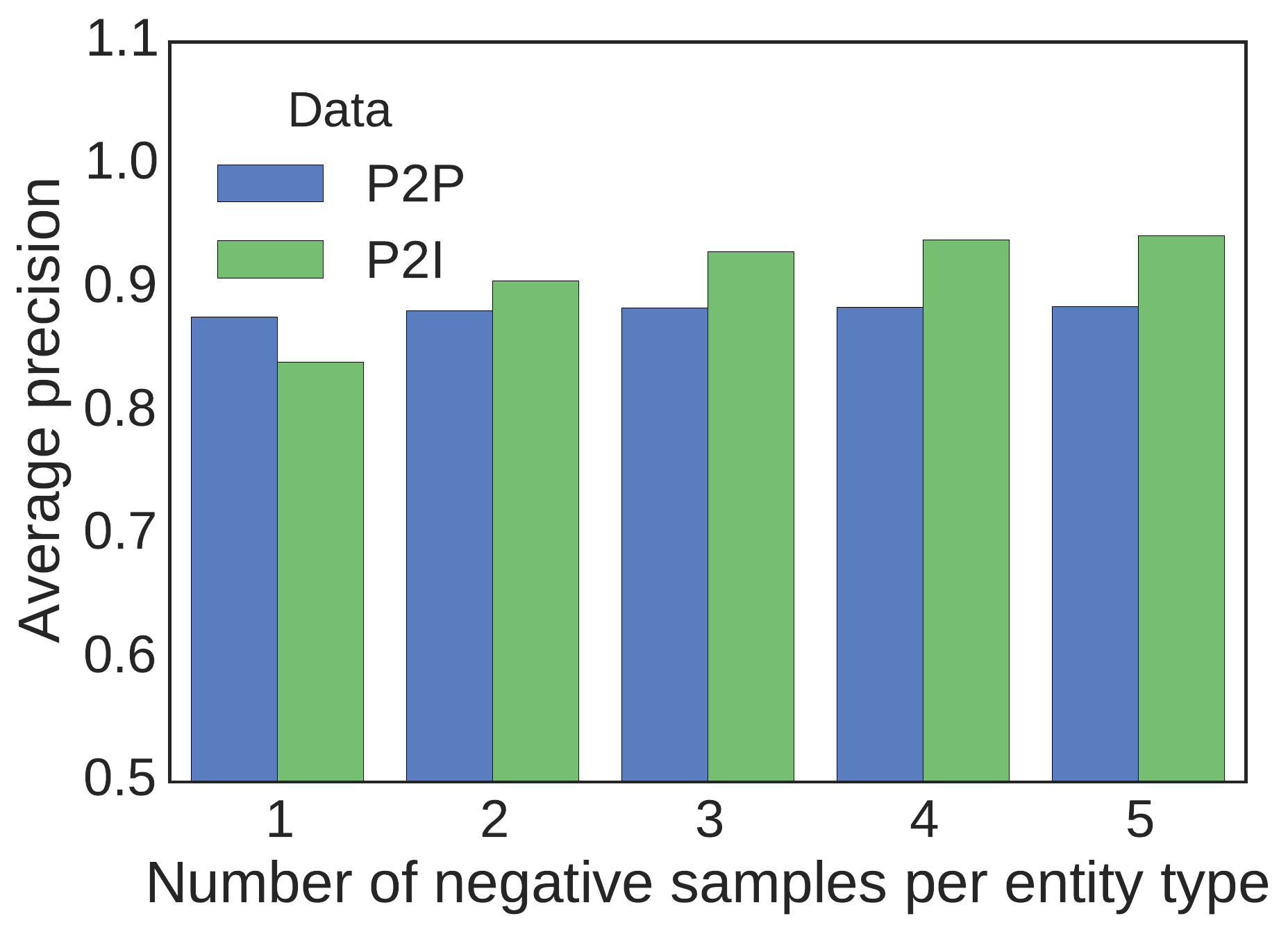, width=0.55\linewidth}
\end{center}
\caption{\label{fig:num_neg_samples} Performance versus number of negative samples drawn per entity type.}
\end{figure}

Figure \ref{fig:num_neg_samples} shows the detection performance versus the number of negative samples drawn per entity type. As we can see, it only requires a reasonable number of negative samples to learn well, though adding more negative samples may marginally improve performances. 

\subsection{A case study for entity embedding}

\begin{figure*}[tp]
\centering
\subfloat[\label{fig:vis_emb_uid}User embeddings.]{\includegraphics[width=2.3in]{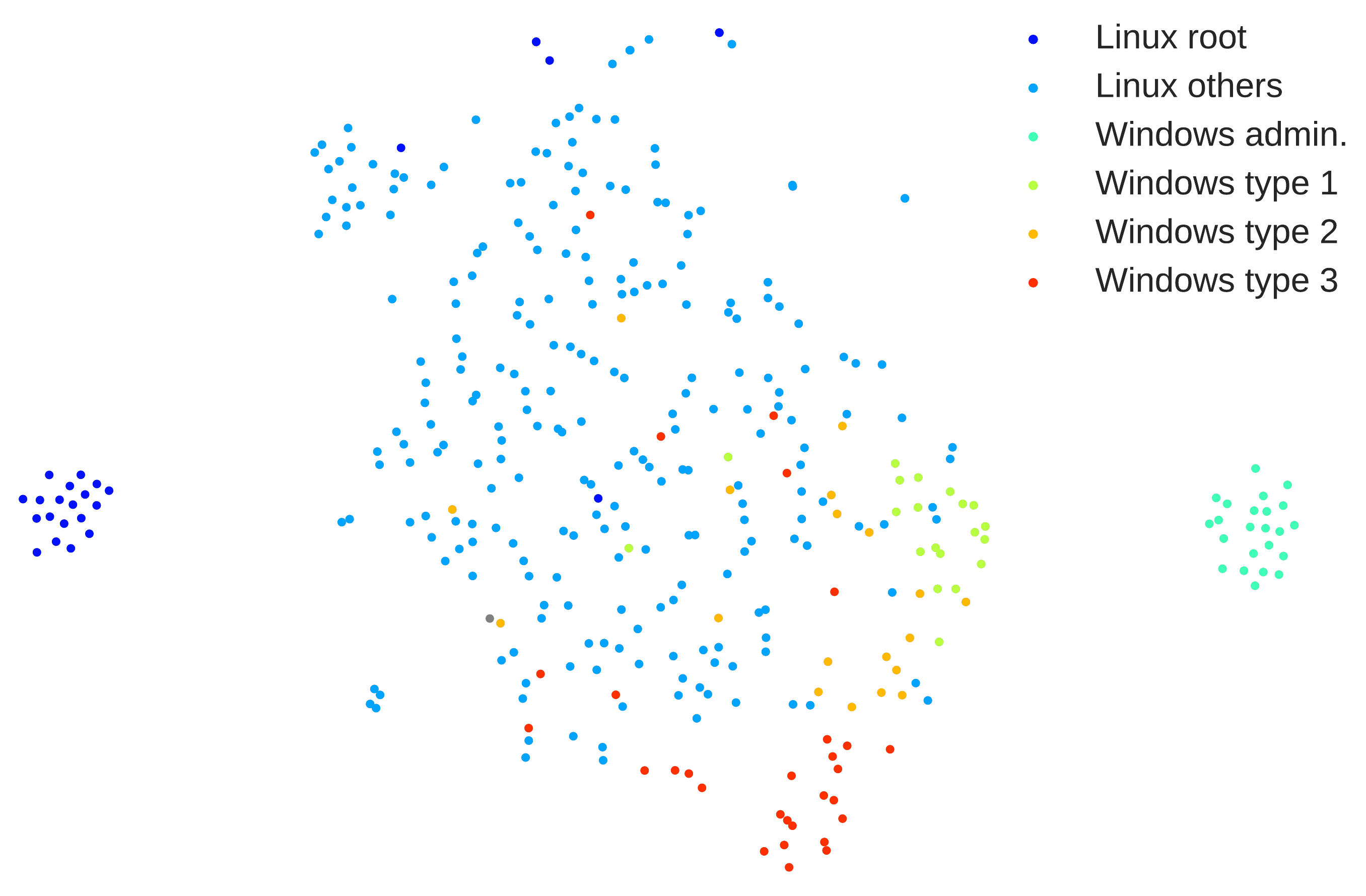}} \hspace{7em}
\subfloat[\label{fig:vis_emb_hour}Hour embeddings.]{\includegraphics[width=2.1in]{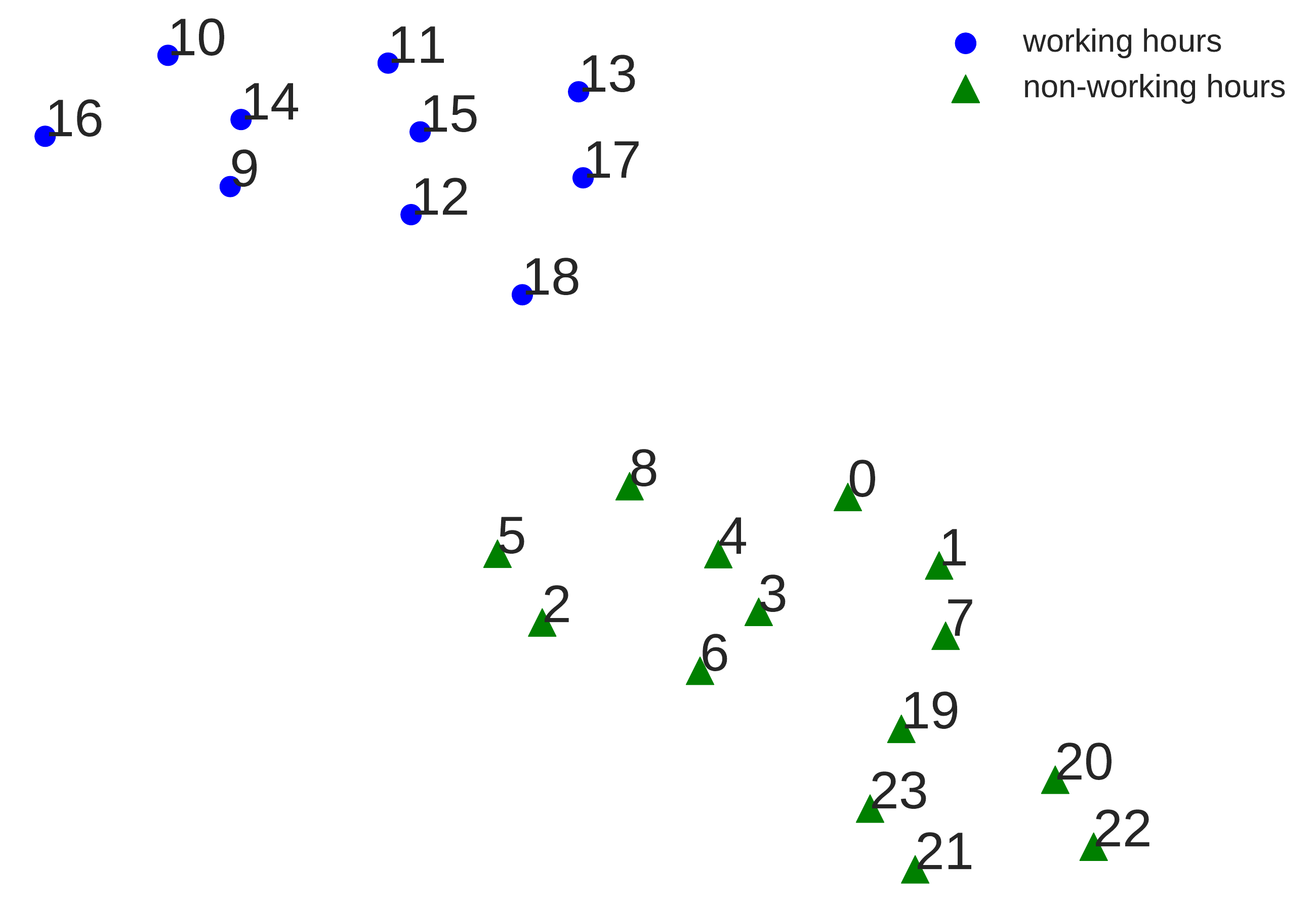}}
\caption{\label{fig:vis_emb}2d visualization of some entity embeddings.}
\end{figure*}

In order to see if the learned embedding is meaningful, we use t-sne \cite{van2008visualizing} to find 2d coordinates of the original entity embeddings. Figure \ref{fig:vis_emb_uid} shows the embedding of users in P2P data. We color each user according to the user type. We find that, in the embedding space, similar types of users are clustered together, indicating they play the same role \cite{ting2016sdm}; and in particular, root users are grouped together and far away from other types of users, reflecting that root users behave very different from other users. Figure \ref{fig:vis_emb_hour} shows the embedding of hours in P2I data. Although not knowing a priori, the \modelname{} model clearly learns the separations of working hours and non-working hours.

Knowing the types of users and differences among hours can be important for detecting abnormal events. The entity embedding learned by the \modelname{} model suggests it can distinguish the semantics/similarities of different entities, thus can help better detect anomalies.

%% file: relatedwork.tex
\section{Related Work}

\subsection{Anomaly Detection}

There are many literatures for anomaly detection, a good summary of the anomaly detection methods can be found in \cite{chandola2009anomaly}. However, most of those work focuses on either numerical data type or supervised settings. 

As for unsupervised categorical anomaly detection, recent work includes \cite{das2007detecting,das2008anomaly,akoglu2012fast}. Most of these methods try to model the regular patterns behind data, and produce abnormal score of data according to some heuristics, such as the compression bits for an event \cite{akoglu2012fast}.

There is some work on applying graph mining methods for anomaly detection in graph \cite{tong2008fast,akoglu2014graph}. However, our setting is different in the sense that, as shown in Section \ref{sec:problem}, when treating categorical events as a network, it is a heterogeneous network \cite{sun2013mining}.

There is also some work on anomaly detection for heterogeneous data \cite{ren2009efficient,das2010multiple},  However, most of them are not suitable for event data due to the lack of distance measure among data points. For example, \cite{das2010multiple} uses LCS to measure distance between two sequences, but will not work for two events.

\subsection{Embedding Methods}

Embedding methods are widely studied in graph/network setting \cite{belkin2001laplacian,tangline}. And more recently, there is some work \cite{bengio2003neural,mikolov2013efficient,mikolov2013distributed} on natural language processing, which tries to embed words into some high dimensional space.

Our work also explores the embedding methods, however, there are some fundamental differences between our method and other embedding methods. Firstly, many of those embedding methods aim to embed pairwise interactions, but they only consider one type of entities. For pairwise interaction of different types of entities, we provide a weighted scheme for distinguishing their importance. Secondly, existing embedding methods cannot be directly applied to predicting abnormal score.

There is some work \cite{agovic2009anomaly} applying graph embedding methods for anomaly detection in numerical data where the distance among data points are easy to compute. However, as far as we know, embedding methods have not explored in anomaly detection applications on categorical event data.

%% file: conclusion.tex
\section{Conclusions}

In this paper, we tackle a challenging problem of anomaly detection on heterogeneous categorical event data. Different from previous work that heavily relies on heuristics, we propose a principled and unified model $\modelname{}$ that directly learns the likelihood of events. The model is instantiated by weighted pairwise interactions among entities that are quantified based on entity embeddings. Using Noise-Contrastive Estimation with ``context-dependent" noise distribution, our model can be learned efficiently regardless of the exponentially large event space. Experimental results on real enterprise surveillance data show that our method can accurately detect abnormal events compared to other state-of-the-art abnormal detection techniques.

As for the future work, it is interesting to consider the temporal correlations among multiple events instead of treating them independently, as many intrusions/attacks can involve a series of events.